\documentclass{article}

\PassOptionsToPackage{numbers}{natbib}

\usepackage[preprint]{neurips_2026}

\usepackage[utf8]{inputenc} 
\usepackage[T1]{fontenc}    
\usepackage{hyperref}       
\usepackage{url}            
\usepackage{booktabs}       
\usepackage{amsfonts}       
\usepackage{nicefrac}       
\usepackage{microtype}      
\usepackage{xcolor}         

\usepackage{cancel}

\usepackage{caption}

\usepackage{multirow}

\usepackage{algorithm}
\usepackage{algpseudocode}

\usepackage{subcaption}

\usepackage{tikz}
\usetikzlibrary{shapes,decorations,arrows,calc,arrows.meta,fit,positioning}
\tikzset{
    -Latex,auto,node distance =1 cm and 1 cm,semithick,
    gray node/.style={circle,fill=gray!20,inner sep=0, minimum size=1.5em}
}

\usepackage{soul} 

\usepackage{amsthm}
\newtheorem{theorem}{Theorem}

\newtheorem{definition}[theorem]{Definition}

\usepackage{algpseudocode}

\usepackage{enumitem}

\usepackage{esvect}

\usepackage{bbm}

\usepackage{mathtools}          

\def \bY{\mathbf{Y}}

\def \bv{\mathbf{v}}
\def \bV{\mathbf{V}}

\def \bU{\mathbf{U}}

\def \bx{\mathbf{x}}
\def \bX{\mathbf{X}}

\def \bF{\mathbf{F}}
\def \bY{\mathbf{Y}}
\def \bZ{\mathbf{Z}}
\def \bz{\mathbf{z}}

\def \bO{\mathbf{O}}
\def \bo{\mathbf{o}}

\def \doI{\text{do}(I)}
\def \doIk{\text{do}(I_k)}
\def \doIj{\text{do}(I_j)}

\def \doEmpty{\text{do}(\emptyset)}
\def \doTzero{\text{do}(T=0)}
\def \doTone{\text{do}(T=1)}

\def \bbI{\mathbb{I}}
\def \bbD{\mathbb{D}}
\def \bbC{\mathbb{C}}
\def \bbR{\mathbb{R}}
\def \bbE{\mathbb{E}}
\def \bbId{\mathbb{I}_{\mathbb{D}}}
\newcommand{\ECpi}[1]{\mathbb{E}_{\fC \sim \pi}\left[#1\right]}

\def \fC{\mathfrak{C}}

\def \cD{\mathcal{D}}

\def \cH{\mathcal{H}}

\newcommand{\ThetaF}[1]{\Theta\left[#1\right]}
\newcommand{\phiF}[1]{\phi\left(#1\right)}

\usepackage[acronym,nohypertypes={acronym,notation}]{glossaries}

\newacronym{scm}{SCM}{structural causal model}
\newacronym{ncm}{NCM}{neural causal model}
\newacronym{po}{PO}{potential outcome}
\newacronym{dag}{DAG}{directed acyclic graph}
\newacronym{mlp}{MLP}{multi-layer perceptron}
\newacronym{cate}{CATE}{conditional average treatment effect}
\newacronym{ite}{ITE}{individual treatment effect}
\newacronym{pma}{PMA}{pooling by multihead
attention}
\newacronym{isab}{ISAB}{induced set attention block}

\usepackage{cleveref}

\title{Computational Identifiability}

\author{%
  Lucius E.J. Bynum\\
  New York University\\
  \texttt{lucius@nyu.edu}\\
  \And
  Rajesh Ranganath\\
  New York University\\
  \texttt{rajeshr@cims.nyu.edu}\\
  \And
  Kyunghyun Cho\\
  New York University\\
  \texttt{kyunghyun.cho@nyu.edu}\\
}

\begin{document}

\maketitle

\begin{abstract}
  Identification conditions describe the computability of a target query or parameter of interest as a function of the type and amount of information available. In causal identification, this information is often expressed in the form of a causal graph, and data are observed or collected for some subset of variables in the graph. Target queries may be for a single effect alone or for a class of effects in a given model. The derivation of an identification algorithm then defines mathematically the process by which the desired causal effect(s) can be uniquely determined, theoretically, in expectation. Identifiability in expectation, or `theoretical identifiability,' generally assumes asymptotic properties, infinite data, or other mathematically idealized conditions. In this paper, we explore a fundamental distinction between this theoretical, idealized notion of identifiability and a proposed alternative that is computation-bound. The framework we propose — `computational identifiability' — is to instead define a finite computational search procedure for an empirical estimator. If this process finds an estimator empirically, within a desired error tolerance, then identifiability is satisfied, \emph{conditional on the specified assumptions of the search (i.e., a prior distribution over the parameters) and conditional on the search procedure itself}. Through several experiments, we demonstrate how this framework allows us to answer fine-grained, practical identification questions, such as identification with small finite samples, with ambiguous graphical criteria, with mixed observational-interventional data, and across counterfactual data and estimands. Code is available at \url{https://github.com/lbynum/metadentify}.
\end{abstract}

\begin{figure*}[h]
\includegraphics[width=\textwidth]{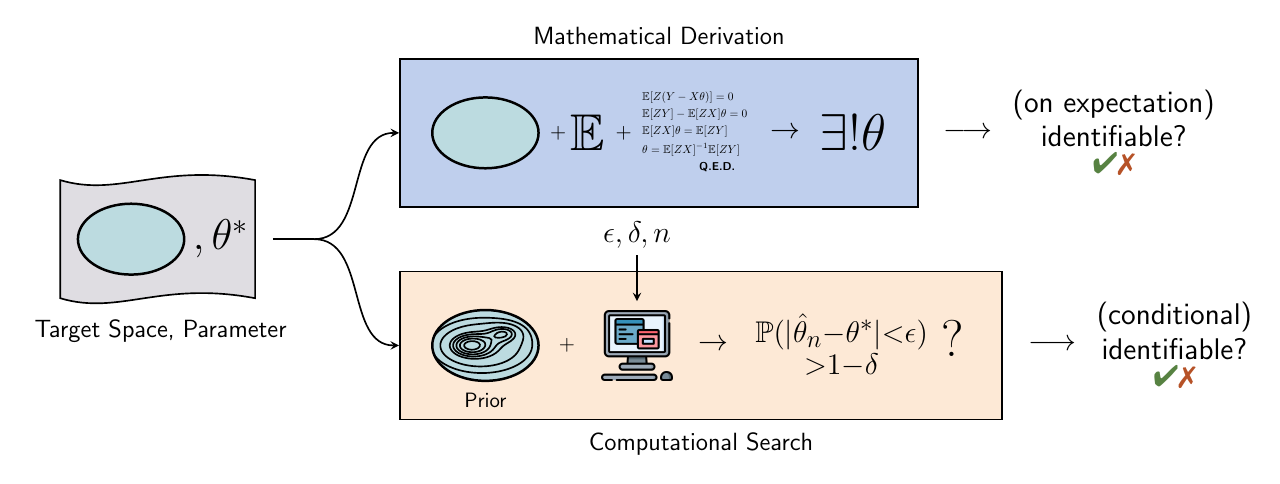}
\caption{Two approaches to identifiability. Mathematical derivation (top) seeks to prove the existence of a unique parameter, analytically in expectation. Computational search (bottom, proposed in this work) instead defines an empirical search procedure for an estimator and defines `computational identifiability' as the successful discovery of an estimator, conditional on a prior over parameters, a hypothesis space over estimators, finite samples, and a desired error tolerance.}\label{fig:figure_1}
\end{figure*}

\section{Introduction}\label{sec:intro}

A critical step in being able to answer a causal or a statistical query from data is determining whether or not there is sufficient information to determine a unique answer. Identifiability addresses this problem, deriving mathematically what is possible to uniquely determine from a collection of distributions that encode assumptions, in expectation. In causal inference, identification algorithms often make use of combinations of causal graphical criteria, probability axioms, matrix or algebraic equations, bounding arguments, and other mathematical tools. This notion of identifiability in expectation, which we term `theoretical identifiability,' generally assumes asymptotic properties, infinite data, or other mathematically idealized conditions.

While theoretical identifiability is a long-solved problem for many model classes in expectation, with sound and complete algorithms like do-calculus \cite{pearl1995probabilistic,shpitser2008complete}, po-calculus \cite{malinsky2019potential}, or $\sigma$-calculus \cite{Correa_Bareinboim_2020}, identifiability itself is far from a solved problem. Determining theoretical identifiability in new settings requires bespoke analysis~\citep{puli2020general, puli2020causal}. Moreover, many important settings involve finite sample sizes, ambiguous graphical criteria, or complex combinations of observational and experimental data that defy the requirements of existing algorithms and known bounds. In such scenarios, theoretical guarantees of identifiability in expectation may offer little guidance on whether a target parameter can actually be estimated. Similarly, many settings that are considered unidentifiable nonparametrically may become identifiable under additional parametric assumptions (such as linearity, non-Gaussianity, or monotonicity \cite{shimizu2006linear, puli2020general}), but exploring these possibilities analytically, especially for nonlinear settings, remains a significant challenge. Novel settings that do not yet have theoretical identifiability results, or where identifiability assumptions are hard to satisfy, leave the search for new identification algorithms and partial identification bounds open and ongoing \cite{raghavan2026causal,ghassami2023partial,tan2024consistency,kline2023recent,lee2020causal}.

In this work, we introduce an alternative notion of identifiability: `computational identifiability,' that instead frames identifiability as a finite computational search (see \Cref{fig:figure_1}). In lieu of mathematical derivation, computational identifiability starts with (1) a meta-prior over the parameters in question: in this case, a distribution over \glspl{scm}; and (2) a hypothesis space of possible estimators. For a given finite sample size $n$, a desired error tolerance $\epsilon$, and a desired confidence bound $\delta$, computational identifiability defines successful identification as the existence of an estimator in the given hypothesis space that is satisfactory, based on the desired error tolerance and desired confidence. In other words, computational identifiability defines a practical and empirical, hence actionable, notion of identifiability.

\textbf{Contributions.} Our main contributions are the following.
\begin{enumerate}
    \item We introduce \textbf{computational identifiability}, a practical, computation-bound notion of identifiability to complement existing theoretical and idealized notions.
    \item We formalize and further generalize the connection between causal effect estimation and meta-learning (\Cref{sec:computational_identifiability}), defining the required components to compute and evaluate computational identifiability (hypothesis space $\cH$, error tolerance $\epsilon$, confidence bound $\delta$, and mixture distributions for custom meta-priors over \glspl{scm} and empirical data).
    \item We empirically demonstrate computational identifiability across a variety of complex and small finite sample settings, including identification with ambiguous graphical criteria (\Cref{sec:oset_results}), with mixed observational-interventional data (\Cref{sec:transportability_results}), and across counterfactual data and estimands (\Cref{sec:counterfactual_results}).
\end{enumerate}

\section{Background and related work} 

\paragraph{Complete identification algorithms.} Existing causal identifiability results and algorithms are sound and complete for certain types of models, such as recursive semi-Markovian \glspl{scm}. Do-calculus \cite{pearl1995probabilistic,pearl2022causal} is a set of rules that systematically answer whether or not an interventional causal query is identifiable based on graphical criteria. Do-calculus has been show to be complete \cite{huang2006pearl,shpitser2006identification} for any query that can be expressed in terms of the do-operator. The ID and IDC algorithms \cite{shpitser2008complete} applying do-calculus are complete identification algorithms for interventional as well as conditional interventional causal queries. To address more complex settings that cannot be expressed using the do-operator (such as path-specific effects and dynamic treatment regimes), po-calculus \cite{malinsky2019potential} further generalizes do-calculus to provide identification for conditional path-specific effects. For counterfactual queries, both unconditional and conditional, corresponding identification algorithms ID* and IDC* are complete \cite{shpitser2008complete}. Beyond hard (atomic) interventions on the target variable of interest, $\sigma$-calculus \cite{Correa_Bareinboim_2020} inference rules provide complete identification with more general (soft) interventions that can be atomic, conditional, or stochastic (see \Cref{sec:interventions}). Analogously, $z$-identifiability \cite{bareinboim2012causal} describes identification of causal effects using surrogate experiments on variables other than the target treatment that are more accessible to manipulation in practice.

\paragraph{Bespoke and partial identification.} In settings where general identification algorithms do not apply or where required identification assumptions are not met, novel conditions and algorithms are developed on a case-by-case basis to either point identify or bound target queries \cite[e.g.,][]{balke1997bounds,horowitz2000nonparametric,cai2008bounds,shalit2017estimating,sachs2023general}. Common example settings with bespoke and/or parametric analysis include the instrumental variable setting \cite{balke1994counterfactual,angrist1996identification,NBERt0259,stock2002testing}, the proximal setting \cite{miao2018identifying,tchetgen2020introduction,ghassami2023partial}, and counterfactual data settings \cite[e.g.,][]{raghavan2026causal}.

\paragraph{Neural Causal Models.} A related line of work on \glspl{ncm} \cite{xia2021causal} also connects identifiability and estimation, but in a fundamentally different manner than our work here. Neural identifiability \cite{xia2022neural} shows that identification within the infinite space of possible \glspl{ncm} is exactly equivalent to identification in \gls{scm}-space.\footnote{Neural identification algorithms (e.g., NeuralID \cite{xia2022neural}) proceed by simultaneously maximizing and minimizing the causal query in \gls{ncm}-space, and if the maximum and minimum query values are equivalent (via hypothesis test), identification is satisfied. The value of the query found during this process can then be used as an estimate.} Unlike neural identifiability, the computational identifiability we propose here does not aim to search the \gls{scm}-space itself. Rather, computational identifiability aims to \emph{directly learn a shortcut from data to estimator}, should such a shortcut exist. For this reason, even empirical approximations of neural identifiability are fundamentally distinct from computational identifiability.

\paragraph{Meta- and in-context learning for causal inference.} A last connected line of research is that on meta-learning, in-context learning, and/or meta-prediction for causal effect estimation \cite{bynum2025bbci,robertsonpfn,balazadehcausalpfn,dhirestimating,ma2025foundation}. Identification and sensitivity analysis in the meta-learning setting has only recently started to be considered \cite{montagna2024demystifying,balazadeh2026iv,javurek2026amortizing}. In this paper, we operationalize computational identifiability as an additional layer of abstraction on top of transformer-based meta-learning, taking causal meta-prediction methods as one sub-component of a larger identifiability pipeline. Importantly, any algorithm --- neural-network-based or not, meta-learning-based or not --- can in principle be incorporated into a computational identifiability search (see \Cref{sec:per_dataset_methods} for an example).

\subsection{Preliminaries}

Let capital letter $X$ denote a random variable, where lowercase letter $X=x$ denotes the value it obtains. Let boldface capital letter $\mathbf{X} = \{X_1, \ldots, X_n\}$ denote a set of random variables, with value $\bX = \bx$. Capital $P_X$ denotes the cumulative distribution function of variable $X$ and lowercase $p_X$ denotes the density (or mass) function. Let $P_{Y \mid X=x}$ denote the conditional distribution of $Y$ given $X=x$ and $P_{Y \mid X}$ denote the collection of $P_{Y \mid X=x}$ for all $x$, i.e., the conditional of $Y$ given $X$. 

\paragraph{Causal models.}
We define a \emph{causal model} as a tuple $\mathcal{M} = (\bV, \bU, \bF)$. In this tuple, $\bV$ is a set of observed variables, $\bU$ a set of unobserved (exogenous) variables, and $\bF$ a set of functions $\{f_i\}_{i=1}^{|\bV|}$ for each $V_i \in \bV$ such that $V_i = f_i(\textsc{PA}_i, U_i)$ where $\textsc{PA}_i \subseteq \bV \setminus \{ V_i\}$ represents the causal parents of $V_i$ and $U_i \subseteq \bU$. A causal model can be pictorially represented as a directed acyclic graph (DAG) with nodes for $\bU, \bV$ and directed edges for $\bF$. A \emph{probabilistic causal model} $(\mathcal{M}, P_\bU)$ adds distribution $P_\bU$ over the unobserved variables. The 4-tuple $\mathfrak{C} = (\bV, \bU, \bF, P_\bU)$ is also commonly referred to as a \emph{\gls{scm}}. An \gls{scm} entails an \emph{observational distribution} $P^{\fC}$ as well as distributions after interventions, which we define as follows.

\paragraph{Interventions.}\label{sec:interventions}
We define an \emph{intervention} $I$ on variable $V_i$ as the substitution of equation $V_i = f_i(\textsc{PA}_i, U_i)$ with a modified mechanism $V_i := \tilde{f}_i(\tilde{\textsc{PA}}_i, \widetilde{U}_i)$, commonly notated $\text{do}(I)$. 
Such `soft' interventions \cite{Eberhardt2007InterventionsAC} allow us to express several types of `regimes' under which data are collected, for example allowing us to describe \emph{transportability}, where we may wish to generalize causal knowledge across different environments with combined observational and/or experimental data \cite{correa2020general}. Soft interventions encompass the \emph{idle} or \emph{null} intervention $\emptyset$, which represents what happens naturally (i.e., $\tilde{f}_i = f_i$); \emph{atomic} interventions, which set $V_i$ to a constant value; \emph{conditional} interventions, which set $V_i$ to a deterministic function of observable parents; and \emph{stochastic} interventions, which set $V_i$ to a probability distribution conditional on a set of observable parents (see \cite{correa2020general} for detailed discussion). A \emph{submodel} $\mathcal{M}_I$ for intervention $I$ is the model $\mathcal{M}$ after intervention $\text{do}(I)$. Given a probabilistic causal model, we can derive from $(\mathcal{M}_I, P_\bU)$ the distribution of any subset of variables following intervention $\text{do}(I)$. We denote an \gls{scm} $\fC$ after intervention $I$ as $\fC^{\doI}$, and the resulting \emph{interventional distribution} as $P^{\fC; \doI}$. Note that $P^{\fC; \doEmpty}$ is equivalent to the observational distribution $P^{\fC}$.

\paragraph{Counterfactuals.} If, instead of using $P_\bU$ in $(\mathcal{M}_\bx, P_\bU)$ to derive an interventional distribution, we specify a distribution over the exogenous variables that is specific to a particular context or individual, the same mechanics that allow us to define interventions allow us to define \emph{counterfactuals} that model alternate possible outcomes after interventions in a specific context. We use an asterisk to denote counterfactual versions $\bV^*$ of variables $\bV$. Counterfactual variable $\bY^*$ given a factual observation $\bz$ and intervention $I$ ($\bY, \bZ \subseteq \bV$) can be computed via a three-step procedure often referred to as `abduction, action, prediction.' Abduction uses observed evidence to obtain $P_{\bU \mid \bz}$ from $P_\bU$. Action performs intervention $I$ to obtain $\mathcal{M}_{I}$. Prediction computes the probability of $\bY^*$ from $(\mathcal{M}_{I}, P_{\bU \mid \bz})$ (or analogously conditioning on $\bZ \in A$ with $P(\bZ \in A) > 0$ in the continuous case rather than on $\bZ=\bz$). We denote the \emph{counterfactual distribution} $P^{\fC \mid \bZ = \bz; \doI}$.

\section{Computational identifiability}\label{sec:computational_identifiability}

In this section, we introduce the key theoretical components that define computational identifiability as a meta-property of a computational search for a causal estimator.
\begin{definition}[Causal query]\label[definition]{def:causal_query}
Let $\bbC$ denote the space of all \glspl{scm} with a fixed set of endogenous variables $\bV$.
We define a \emph{causal query} $\Theta[\cdot]$ as a functional taking as input an \gls{scm} $\fC \in \bbC$ as well as a query point $\bX = \bx$ for any $\bX \subseteq \bV$. More formally, let $\Omega_{\bX}$ denote the sample space for any subset $\bX \subseteq \bV$ and define $\Omega = \bigsqcup_{\bX \subseteq \bV} \Omega_{\bX}$. Then $\Theta: \bbC \times \Omega \rightarrow \bbR$.
\end{definition}
For example, average treatment effects such as the population average treatment effect (PATE) and conditional average treatment effect (CATE) can be defined as expectations of interventional distributions specific to a given \gls{scm} $\fC$ that either make use of or ignore a given observation $\bx$. Specifically, given \gls{scm} $\fC = (\bV, \bU, \bF, P_\bU)$ with a binary treatment variable $T \in \bV$ and outcome variable $Y \in \bV$, the PATE can be written as 
$\Theta_{\text{PATE}}\left[\fC, \cdot \right] := \mathbb{E}_{p_{Y}^{\fC; \doTone}}[Y] - \mathbb{E}_{p_{Y}^{\fC; \doTzero}}[Y]$, where the second argument is ignored.
Adding a condition $\bX=\bx$ for conditioning variables $\bX \subseteq \bV \setminus \{Y, T\}$, the \text{CATE} can be written as 
$\Theta_{\text{CATE}}\left[\fC, \bx \right] := \mathbb{E}_{p_{Y \mid \bX=\bx}^{\fC; \doTone}}[Y] - \mathbb{E}_{p_{Y \mid \bX=\bx}^{\fC; \doTzero}}[Y].$
Common estimands comprised of counterfactual rather than interventional distributions include individual treatment effects (ITEs) and sample average treatment effects (SATEs). The ITE represents the counterfactual difference in outcomes specific to a given unit with observed features $\bV=\bv$ (including realized $T$ and $Y$ values): 
$\Theta_{\text{ITE}}\left[\fC, \bv \right] := \mathbb{E}_{p_Y^{\fC \mid \bV = \bv; \doTone}}[Y] - \mathbb{E}_{p_Y^{\fC \mid \bV = \bv; \doTzero}}[Y].$ 
Building on the ITE, the SATE represents the average ITE across a given set of observed units, which can be viewed as a composition of a set of causal queries. For example, given a dataset $\cD = \{\bv_i\}_{i=1}^{n}$, the SATE can be written as 
$\frac{1}{n} \sum_{i=1}^{n} \Theta_{\text{ITE}}\left[\fC, \bv_i \right].$

\begin{definition}[Causal mixture distribution]\label[definition]{def:causal_mixture}
Let $\bbI = \{\emptyset, I_1, I_2, ..., I_K\}$ represent a set of interventions indexed by $k$, where $\emptyset$ represents the null-intervention (doing nothing). Let $\bbId \subseteq \bbI$ represent the set of interventions for which empirical data are collected. For example, if empirical data are all observational, $\bbId = \{\emptyset\}$. If in addition to observational data, data are collected from a randomized control trial conducted with a treatment intervention $I_{treated}$, a control intervention $I_{control}$, and a placebo intervention $I_{placebo}$, then $\bbId = \{\emptyset, I_{treated}, I_{control}, I_{placebo}\}$.
Let $\bbD = \{\cD_j\}_{j=1}^{|\bbId|}$ represent the corresponding set of empirical datasets (multisets). We also index $\bbId$ with $j$ such that $\cD_j$ corresponds to intervention $I_j \in \bbId$. The observed variables (observed in each dataset $\cD_j \in \bbD$) are denoted $\bO$.
We then define a \emph{causal mixture distribution} given \gls{scm} $\fC$ over endogenous variables $\bV$ where $\bO \subseteq \bV$ as a weighted sum of empirical distributions and \gls{scm}-based observational, interventional, and counterfactual distributions:
\begin{align*}
p^{\fC}_{\bO, mix}(\bo) &:= \sum_{I_j \in \bbId} \sum_{\bo_i \in \cD_j} \left( \smash{\underbrace{\alpha^{emp}_{ij} \delta(\bo - \bo_i)}_{\substack{\textrm{item-weighted empirical}\\\textrm{distribution collected}\\\textrm{under intervention $I_j$}}}} + \sum_{I_k \in \bbI} \alpha^{cf}_{ijk} \smash{\underbrace{p_{\bO}^{\fC^{\doIj}|\bO=\bo_i; \doIk}(\bo)}_{\substack{\textrm{counterfactual distribution under}\\\textrm{\gls{scm} $\fC^{\doIj}$ given evidence $\bo_i$}\\\textrm{and hypothetical intervention $I_k$}}}}\  \right) \\ \\
&+ \sum_{I_k \in \bbI} \alpha^{int}_k  \vphantom{p_{\bO}^{\fC; \doIk}}\smash{\underbrace{p_{\bO}^{\fC; \doIk}(\bo)}_{\substack{\textrm{interventional} \\ \textrm{distribution for} \\ \textrm{$I_k$ under \gls{scm} $\fC$}}}}\\
\end{align*}
with $\sum_{ij} \alpha^{emp}_{ij} + \sum_{ijk} \alpha^{cf}_{ijk} + \sum_{k} \alpha^{int}_{k} = 1$, $\alpha^{emp}_{ij} \geq 0$, $\alpha^{cf}_{ijk} \geq 0$, $\alpha^{int}_{k} \geq 0$, and Dirac delta $\delta$.\footnote{For valid densities, we consider $p^{\fC}_{\bO, mix}$ with respect to a base measure that dominates both the Lebesgue measure and the counting measure of the empirical data.}
\end{definition}

Traditional `theoretical' definitions of causal identifiability are about being able to uniquely determine a causal query from a given (typically observational) distribution or set of distributions. We define this in our context as follows.
\begin{definition}[Theoretical identifiability]\label[definition]{def:theoretical_identifiability} Given a space $\Omega_{\fC}$ over \glspl{scm} $\fC$ with endogenous variables $\bV$ and a subset of observed variables $\bO \subseteq \bV$, we say a causal query $\Theta\left[\fC, \bx \right]$ with query point $\bX = \bx$ for $\bX \subseteq \bV$ is \emph{theoretically identifiable} from $\bO$ if
for any two \glspl{scm} $\fC_1, \fC_2 \in \Omega_{\fC}$, $\Theta\left[\fC_1, \bx \right] = \Theta\left[\fC_2, \bx \right]$ whenever $P_{\bO}^{\fC_1} = P_{\bO}^{\fC_2}$.\footnote{We have written this definition using observational distribution $P_{\bO}^{\fC}$ by default, but for theoretical identifiability from a different class of distribution, the analogous equality would be between whichever class of distributions is available.} A consequence is that $\Theta$ can be written as a functional of $P_{\bO}^{\fC}$.
\end{definition}
Given our setup, we can instead define a new notion of identifiability that is conditional on a particular prior $\pi$ over \glspl{scm} and a hypothesis space $\cH$ over functions derived from a causal mixture distribution.
\begin{definition}[Computational identifiability]\label[definition]{def:computational_identifiability} Given a prior distribution over \glspl{scm} $\fC \sim \pi$, a causal mixture distribution $p^{\fC}_{\bO,mix}$ over observed variables $\bO$ with sample space $\Omega_{\bO}$, and causal query $\ThetaF{\fC, \bx}$ with query point $\bX = \bx$ ($\bX \subseteq \bV$) with sample space $\Omega_{\bX}$, we say $\Theta$ is \emph{$\epsilon$-$\delta$-identifiable conditional on prior $\pi$ and hypothesis space $\cH$} if there exists a function $\phi \in \cH$ where $\phi: \Omega_{\bO} \times \Omega_{\bX} \mapsto \bbR$ such that $\ECpi{\bbE_{\bO \sim p^{\fC}_{\bO,mix}} \left[\mathbf{1}\left(\left|\ThetaF{\fC, \bx} - \phiF{\bO, \bx}\right| \leq \epsilon \right)\right]} \geq 1 - \delta$. In other words, the query can be approximated within an $\epsilon$ margin of error by $\phi$ with probability at least $1 - \delta$.
\end{definition}

Although causal query $\Theta\left[\fC, \bx \right]$ is evaluated for \gls{scm} $\fC$ pointwise at query point $\bX=\bx$, in practice, our computational search for a causal estimator will involve a distribution not only over observed variables $\bO$ but also over query points $\bX$. We formalize this process via the causal query distribution.

\begin{definition}[Causal query distribution]\label[definition]{def:query_distribution}
Given \gls{scm} $\fC$ over endogenous variables $\bV$, observation $\bO=\bo$ where $\bO \subseteq \bV$, and causal query $\Theta\left[\fC, \bx \right]$ taking as input $\fC$ and query point $\bX=\bx$ where $\bX \subseteq \bV$,
we define causal query distribution
$p_{query}(\bx \mid \fC, \bo)$ as a distribution over $\bX$ conditional on $\fC$ and $\bo$.
\end{definition}

\Cref{def:query_distribution} covers several cases. For a fixed query point $\bX=\bx'$, we would have $p_{query}(\bx \mid \fC, \bo) = \delta(\bx - \bx')$. For query points sampled from the marginal distribution of $\bX$ entailed by $\fC$, we would have $p_{query}(\bx \mid \fC, \bo) = p_{\bX}^{\fC}$. For an empirical query like the ITE evaluated at observation $\bO=\bo$, we might instead have $p_{query}(\bx \mid \fC, \bo) = \delta(\bx - \bo)$ (if, e.g., $\bO$ and $\bX$ represent the same set of variables).

Recent meta-learning and in-context learning approaches to causal effect estimation allow us to amortize the learning of posteriors over causal estimators, based on a specified input prior over \glspl{scm} \cite{bynum2025bbci,robertsonpfn,balazadehcausalpfn}. Therefore, an analogous meta-learning approach would allow us to learn a \emph{posterior over such functionals $\phiF{\bO, \bx}$} and in turn estimate the \emph{probability of identifiability}:
\begin{equation}\label{eq:prob_identifiability}
\textrm{P(identifiability)} = \ECpi{\bbE_{\bO \sim p^{\fC}_{\bO,mix}} \left[ \bbE_{\bX \sim p_{query}} \left[\mathbf{1}\left(\left|\ThetaF{\fC, \bX} - \phiF{\bO, \bX}\right| \leq \epsilon \right) \right]\right]}.
\end{equation}
Intuitively, computational identifiability (e.g., estimating \Cref{eq:prob_identifiability} empirically) combines a few distinct additions to and/or relaxations of theoretical identifiability:
\begin{itemize}[noitemsep]
    \item \textbf{Prior-conditioning:} is the distribution of the target parameter, conditional on the information we have, sufficiently concentrated around a unique value?
    \item \textbf{Hypothesis-conditioning/learnability:} does an estimator exist within our chosen hypothesis space that could predict the target parameter, and can we learn such an estimator empirically in practice? 
    \item \textbf{Approximation/partial identification:} rather than perfect prediction of the target parameter, can we predict the target parameter within an acceptable error tolerance?
\end{itemize}

\subsection{Meta-learning an estimator $\phi$}

In order to find an estimator $\phi$ (or see if it is possible to find an estimator) in a given setting of interest, we have to learn a mapping from observations and query points $(\bO=\bo, \bX=\bx)$ to causal query values $\Theta=\theta$. After choosing a hypothesis space $\cH$, one way to find such an estimator is to meta-learn the desired mapping in a supervised manner from many examples \cite{bynum2025bbci}. To do this in practice, we first define a joint distribution over causal mixture samples and causal query values, from which we can stream training data.

\begin{definition}[Joint causal-query-mixture distribution]\label[definition]{def:joint_causal_query_mixture}
Let $\pi$ represent a distribution over \glspl{scm} $\fC \sim \pi$ where all \glspl{scm} share the same fixed set of endogenous variables $\bV$, and let $\Theta$ represent a causal query of interest with query point $\bX=\bx$ where $\bX \subseteq \bV$. Assume causal mixture distribution $p^{\fC}_{\bO, mix}(\bo)$ and causal query distribution $p_{query}(\bx \mid \fC, \bo)$ are specified for each \gls{scm} $\fC$. We can then define a \emph{joint causal-query-mixture distribution} over observations $\bO = \bo$ and realized causal query values $\Theta = \theta$ as 
$$p_{joint}(\bo, \bx, \theta) = \ECpi{p^{\fC}_{\bO, mix}(\bo) \cdot p_{query}(\bx \mid \fC, \bo) \cdot \delta\left(\theta - \ThetaF{\fC, \bx} \right)}$$
allowing for joint samples of datapoints, query points, and causal query values $\{(\bo, \bx, \theta)\}$.
\end{definition}

\subsubsection{Architectures suitable for meta-learning of $\phi$}\label{sec:architectures}

To meta-learn $\phi$, we consider in this work a couple variants of Conditional Neural Process (CNP) style architectures \cite{garnelo2018conditional,garnelo2018neural}. Neural Processes (NPs) are a family of flexible meta-learning models that operate on set-valued inputs and produce predictions, with uncertainty estimates, over outputs at arbitrary locations \cite{pmlr-v267-ashman25a}. In this section, we briefly describe NPs and map them to our setting as a representative model class that we will use for our analysis in this paper. However, we emphasize that the component of our pipeline that performs the meta-learning of an effect estimation algorithm could alternatively leverage other types of models, such as ST-based architectures \cite{bynum2025bbci,zhang2022set}, or tabular foundation model (TFM) style architectures \cite{qu2026tabiclv2,balazadehcausalpfn,robertsonpfn,ma2025foundation}.

\citet[Appx. A]{pmlr-v267-ashman25a} provide the following unifying construction of the many variants of CNPs as a sequence of three components. Given a context set $\mathcal{D}_c = \{\mathbf{x}_c, \mathbf{y}_c\}$ and a target set $\mathcal{D}_t = \{\mathbf{x}_t, \mathbf{y}_t\}$, the \emph{encoder} $e: \mathcal{X} \times \mathcal{Y} \rightarrow \mathcal{Z}$ encodes each $(\mathbf{x}_{c,n}, \mathbf{y}_{c,n}) \in \mathcal{D}_c$ into a latent representation $\mathbf{z}_{c,n} \in \mathcal{Z}$, the \emph{processor} $\rho: \left( \bigcup_{n=0}^{\infty} \mathcal{Z}^n \right) \times \mathcal{X} \rightarrow \mathcal{Z}$ then processes the embedded context $e(\mathcal{D}_c)$ along with the target input $\mathbf{x}_t$ to obtain target-dependent output $\mathbf{z}_t \in \mathcal{Z}$, and the \emph{decoder} $d:\mathcal{Z} \rightarrow \mathcal{P}_{\mathcal{Y}}$ maps the target input to the predictive distribution over the output at that target location (where $\mathcal{P}_{\mathcal{Y}}$ is the set of distributions over $\mathcal{Y}$). 

In our case, we define a context set $\mathcal{D}_c = \{\mathbf{o}_{c,n}, \mathbf{s}_{c,n}\}_n$ of observations $\mathbf{o}_{c,n}$ and \emph{source labels} $\mathbf{s}_{c,n}$ for each observation $n$. Source labels can be, e.g., indicators for `observational', `interventional,' `counterfactual,' etc., corresponding to each $I_k \in \mathbb{I}$. We then define the target set $\mathcal{D}_t = \{\mathbf{x}_{t,m}, \boldsymbol{\theta}_{t,m}\}_m$ consisting of $m$ query points $\mathbf{x}_{t,m}$ and corresponding realized causal query values $\boldsymbol{\theta}_{t,m}$. Context and target sets $\{(\mathcal{D}_c, \mathcal{D}_t)\}$ are generated from a joint causal-query-mixture distribution directly following \Cref{def:joint_causal_query_mixture}. Following the CNP setup, the model then performs the following operations (see also \cite[Fig. 4]{pmlr-v267-ashman25a}).
\begin{equation*}
\mathcal{D}_c = \{\mathbf{o}_{c,n}, \mathbf{s}_{c,n}\}_n \xrightarrow[]{\text{Encode, } e(\cdot)} \{\mathbf{z}_{c,n}\}_n \xrightarrow[]{\text{Process, } \rho(\cdot, \mathbf{x}_t)} \mathbf{z}_t \xrightarrow[]{\text{Decode, } d(\cdot)} p(\boldsymbol{\theta}_t \mid \mathcal{D}_c, \mathbf{x}_t)
\end{equation*}
In the experiments in this paper, we make use of two CNP variants based on the above setup. The first is a standard mean-pooled CNP, and the second is an attention-based variant utilizing Set Transformers \cite{lee2019set} (a TNP variant). C/TNPs often assume a Gaussian distribution over the target variable and thus output the target mean and variance (trained via, e.g., Gaussian Negative Log Likelihood Loss). In our causal inference setting, with treatment effect distributions that can be skewed, heavy-tailed, or multi-modal, we instead opt to implement quantile regression similar to the Conditional Quantile Neural Process (CQNP) \cite{mohseni2023adaptive}. In this vein, we take the desired quantile level $\tau \sim \mathcal{U}(0, 1)$ as an input during the forward pass. $\tau$ is embedded (via \acrshort{mlp} in the CNP or a Cosine Embedding Network in the TNP) and concatenated with the aggregated context before decoding. We train via Huber loss over randomly sampled quantiles for each batch. This allows the decoder to act as a continuous quantile function $Q(\boldsymbol{\theta}_t \mid \mathcal{D}_c, \mathbf{x}_t, \tau)$. We refer to the two architectures we use in our experiments in this work as Q-CNP and Q-TNP, if relevant, and otherwise just say `the meta-model.'

\textbf{PATE vs. CATE/ITE meta-estimation.} Architecturally, the distinction between PATE and CATE/ITE meta-estimation reduces to the routing of the query points $\mathbf{x}_t$. For population-level estimands like PATE, the decoder exclusively uses the global context, whereas for individual-level estimands like CATE/ITE, the meta-model conditions on $\mathbf{x}_t$, processing query features with the global context either via direct concatenation (Q-CNP) or cross-attention (Q-TNP). 

\subsection{Posterior estimates of the probability of identifiability}
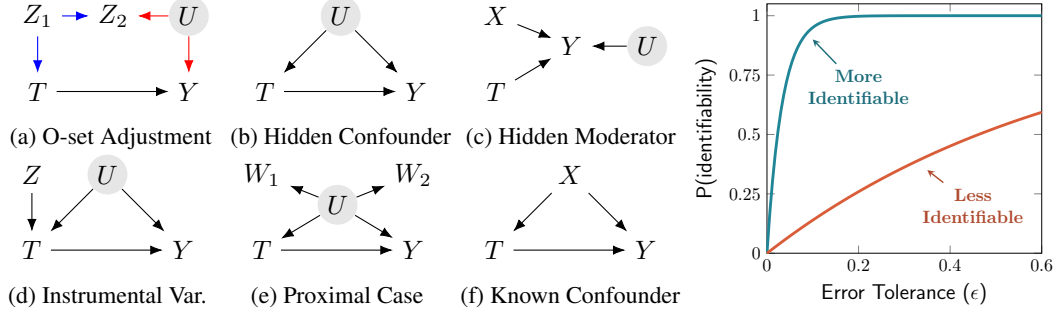
\begin{figure}
\begin{minipage}[b]{0.65\textwidth}
\begin{subfigure}{0.33\textwidth}
    \centering
    \begin{tikzpicture}
        \node (t) at (0, 0) {$T$};
        \node (z1) at (0, 1) {$Z_1$};
        \node (z2) at (1, 1) {$Z_2$};
        \node[gray node] (u) at (2, 1) {$U$};
        \node (y) at (2, 0) {$Y$};
    
        \path[blue] (z1) edge (z2);
        \path[blue] (z1) edge (t);
        \path (t) edge (y);
        \path[red] (u) edge (z2);
        \path[red] (u) edge (y);
    \end{tikzpicture}
    \caption{O-set Adjustment}\label{fig:oset_dag}
\end{subfigure}
\begin{subfigure}{0.32\textwidth}
    \centering
    \begin{tikzpicture}
        \node (t) at (0, 0) {$T$};
        \node[gray node] (u) at (1, 1) {$U$};
        \node (y) at (2, 0) {$Y$};
    
        \path (t) edge (y);
        \path (u) edge (t);
        \path (u) edge (y);
    \end{tikzpicture}
    \caption{Hidden Confounder}\label{fig:hidden_confounder_dag}
\end{subfigure}
\begin{subfigure}{0.33\textwidth}
    \centering
    \begin{tikzpicture}    
        \node (t) at (0, 0) {$T$};
        \node (x) at (0, 1) {$X$};
        \node[gray node] (u) at (2, 0.6) {$U$};
        \node (y) at (1, 0.6) {$Y$};
    
        \path (t) edge (y);
        \path (x) edge (y);
        \path (u) edge (y);
    \end{tikzpicture}
    \caption{Hidden Moderator}\label{fig:hidden_moderator_dag}
\end{subfigure}
\begin{subfigure}{0.32\textwidth}
    \centering
    \begin{tikzpicture}
        \node (t) at (0, 0) {$T$};
        \node (z) at (0, 1) {$Z$};
        \node[gray node] (u) at (1, 1) {$U$};
        \node (y) at (2, 0) {$Y$};
    
        \path (z) edge (t);
        \path (t) edge (y);
        \path (u) edge (t);
        \path (u) edge (y);
    \end{tikzpicture}
    \caption{Instrumental Var.}\label{fig:instrument_dag}
\end{subfigure}
\begin{subfigure}{0.33\textwidth}
    \centering
    \begin{tikzpicture}
        \node (t) at (0, 0) {$T$};
        \node (w1) at (0, 1) {$W_1$};
        \node (w2) at (2, 1) {$W_2$};
        \node[gray node] (u) at (1, 0.6) {$U$};
        \node (y) at (2, 0) {$Y$};
    
        \path (u) edge (w1);
        \path (u) edge (w2);
        \path (t) edge (y);
        \path (u) edge (t);
        \path (u) edge (y);
    \end{tikzpicture}
    \caption{Proximal Case}\label{fig:proximal_dag}
\end{subfigure}
\begin{subfigure}{0.33\textwidth}
    \centering
    \begin{tikzpicture}
        \node (t) at (0, 0) {$T$};
        \node (x) at (1, 1) {$X$};
        \node (y) at (2, 0) {$Y$};
    
        \path (t) edge (y);
        \path (x) edge (t);
        \path (x) edge (y);
    \end{tikzpicture}
    \caption{Known Confounder}\label{fig:known_confounder_dag}
\end{subfigure}
\end{minipage}
\begin{subfigure}[t]{0.35\textwidth}
    \centering
    \includegraphics[width=\textwidth]{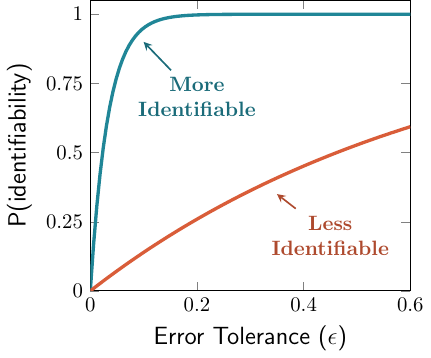}
\end{subfigure}
\caption{\textbf{(a)}-\textbf{(f)} DAGs for each experimental setting we consider. \textbf{(Right)} Diagram showing \emph{computational identifiability curves}, visualizing \Cref{def:computational_identifiability} across a range of possible $\epsilon$ values. The empirical (or posterior) probability of identifiability can be read at a given desired error tolerance.}\label{fig:dags_and_id_curve_diagram}
\end{figure}

\begin{algorithm}
\caption{Estimate $\text{P(identifiability)}$ at $\epsilon$ via posterior uncertainty}\label{alg:posterior_p_identifiability}
\begin{algorithmic}[1]
\footnotesize
\Require Test set $\mathcal{D}_{\text{test}}$, trained quantile function $\widehat{Q}$, dense quantile grid $\mathcal{T} = \{\tau_1, \dots, \tau_K\}$, tolerance $\epsilon$
\For{$(\mathbf{o}, \mathbf{x}) \in \mathcal{D}_{\text{test}}$}
    \State $\phi \gets \widehat{Q}_{0.5}(\mathbf{o}, \mathbf{x})$
    \State $\widehat{P}(\mathbf{o}, \mathbf{x}) \gets \frac{1}{K} \sum_{k=1}^K \mathbf{1}\left( \widehat{Q}_{\tau_k}(\mathbf{o}, \mathbf{x}) \in [\phi - \epsilon, \phi + \epsilon] \right)$
    \Comment{$\approx \int_0^1 \mathbf{1}\left( \left| \widehat{Q}_\tau(\mathbf{o}, \mathbf{x}) - \phi(\mathbf{o}, \mathbf{x}) \right| \le \epsilon \right) d\tau$}
\EndFor
\State $\widehat{P}_{\text{avg}} \gets \frac{1}{|\mathcal{D}_{\text{test}}|} \sum_{(\mathbf{o}, \mathbf{x}) \in \mathcal{D}_{\text{test}}} \widehat{P}(\mathbf{o}, \mathbf{x})$
\Comment{$\approx \ECpi{\bbE_{\bO \sim p^{\fC}_{\bO,mix}} \left[ \bbE_{\bX \sim p_{query}} \left[\widehat{P}(\mathbf{O}, \mathbf{X}) \right]\right]}$}
\State \Return $\widehat{P}_{\text{avg}}$
\end{algorithmic}
\end{algorithm}
A benefit of estimation with uncertainty in our context is the ability to, given a trained model, directly estimate the probability of identifiability for any new data at test-time. \Cref{alg:posterior_p_identifiability} describes this process in the case of quantile regression (applicable to Q-CNP and Q-TNP), but analogous procedures would work for any model supporting posterior uncertainty. A more direct way to estimate computational identifiability than \Cref{alg:posterior_p_identifiability} is to directly compute the rate of predictions exceeding the desired error tolerance empirically over a test set. We do both in this paper. The benefit of theoretical estimates such as in \Cref{alg:posterior_p_identifiability} is that we can in principle estimate computational identifiability for new data without ground truth effects. We leave a full characterization of this process to future work, but demonstrate how \Cref{alg:posterior_p_identifiability} lines up with empirical estimates for in-meta-distribution samples in our experiments in \Cref{fig:ite_curves}. \Cref{fig:dags_and_id_curve_diagram} (right) explains an example diagram of computational identifiability (the probability of identifiability) of the kind we compute throughout our experiments.

\section{Experiments}\label{sec:experiments}

In this section, we empirically demonstrate the computational identifiability framework across a variety of complex and small finite sample settings, including identification with ambiguous graphical criteria (\Cref{sec:oset_results}), with mixed observational-interventional data (\Cref{sec:transportability_results}), and across counterfactual data and estimands (\Cref{sec:counterfactual_results}). \Cref{sec:generalizing_other_works} additionally includes experiments that illustrate how computational identifiability further generalizes other works that explore causal meta-prediction. Architectures, hyperparameter settings, and training details can be found for each experiment in \Cref{tab:meta_model_lookup} in \Cref{sec:experiment_details}.

\subsection{Optimal adjustment set selection}\label{sec:oset_results}

One use-case for computational identifiability is to distinguish between settings that are --- from a theoretical identifiability perspective --- all equally identifiable. A common setting for this is the problem of determining an \emph{optimal valid adjustment set (O-set)} \cite{henckel2022graphical} from among a set of valid covariate adjustment sets, defined in this case as the valid adjustment set that provides optimal asymptotic variance for estimation. 

Works such as \cite{henckel2022graphical} provide graphical criteria for determining the O-set in linear causal models, extended to the non-parametric case in \cite{rotnitzky2020oset}. \citet{rotnitzky2020oset} also discuss several cases where an optimal adjustment set does not exist according to graphical criteria. One such example is the \gls{dag} shown \Cref{fig:oset_dag}, with treatment $T$, outcome $Y$, and unobserved variable $U$. There are three valid adjustment sets: $\mathbf{Z}^0=\emptyset$, $\mathbf{Z}^1 = \{Z_1\}$, and $\mathbf{Z}^{12} = \{Z_1, Z_2\}$. \citet{rotnitzky2020oset} show that $\mathbf{Z}^0$ is uniformly better than $\mathbf{Z}^1$, but that determining which adjustment set is optimal between $\mathbf{Z}^0$ and $\mathbf{Z}^{12}$ would depend on the structural equations: $\mathbf{Z}^0$ is better than $\mathbf{Z}^{12}$ if the blue edges encode strong associations and the red edges are weak, while $\mathbf{Z}^{12}$ is better than $\mathbf{Z}^0$ if instead the red edges are strong and the blue edges are weak. \citet[Ex. 5]{rotnitzky2020oset} demonstrate this by showing the existence of two parameter settings of \Cref{eq:oset_equations} --- one where $\mathbf{Z}^0$ is optimal and one where $\mathbf{Z}^{12}$ is optimal.
\begin{equation}\label{eq:oset_equations}
\begin{aligned}
U &\sim \text{Bern}(p_u) \\
Z_1 &\sim \text{Bern}(p_{z_1}) \\
\end{aligned}
\quad \quad
\begin{aligned}
A &\sim \text{Bern}(p_{1}^{A} Z_1 + p_{0}^{A} \lnot Z_1) \\
Z_2 &\sim \text{Bern}(p_{11}^{Z_2} Z_1 U + p_{10}^{Z_2} Z_1 \lnot U + p_{01}^{Z_2} \lnot Z_1 U + p_{00}^{Z_2} \lnot Z_1 \lnot U) \\
Y &\sim \text{Bern}(p_{11}^{Y} U A + p_{10}^{Y} U \lnot A + p_{01}^{Y} \lnot U A + p_{00}^{Y} \lnot U \lnot A)
\end{aligned}
\end{equation}
But which adjustment set would be optimal in general, if, say, we had a prior over the parameters $\pi(p_j)$ and an observed dataset of size $n=1000$? Computational identifiability can answer this question directly: given a prior and hypothesis space of estimators, we simply meta-train an estimator and inspect which adjustment set leads to the lowest test error. 

\Cref{tab:oset_table} shows exactly this comparison, showing test RMSE for the meta-model. In the first row, the prior over structural equations corresponds to \Cref{eq:oset_equations} with parameters $p_j \sim \mathcal{U}(0, 1)$ for all $j$. In this case, estimation is easy, there is no noise, and all models have very low RMSE; however, we could still say $\mathbf{Z}^{12}$ is the \emph{conditional O-set}. But what about in a case where estimation is difficult? The second row in \Cref{tab:oset_table} shows the case where instead the prior over structural equations is random \glspl{mlp} (each with 2 layers) with Gaussian noise: $U \sim \mathcal{N}(0, 1)$, $Z_1 \sim \mathcal{N}(0, 1)$, $Z_2 = \text{RandomMLP}(Z_1, U) + \mathcal{N}(0, 1)$, $A = \text{RandomMLP}(Z_1) + \mathcal{N}(0, 1)$, and $Y = \text{RandomMLP}(A, U) + \mathcal{N}(0, 1)$. In this case, following the same empirical process, we can determine the conditional O-set is instead $\mathbf{Z}^{0}$.

\begin{table}
\caption{Meta-model performance results for optimal adjustment set selection.}
\label{tab:oset_table}
\centering
\begin{tabular}{llll}
\toprule
\multirow{2}{*}{\textbf{SCM Function Class}} & \multicolumn{3}{c}{\textbf{RMSE}}\\\cmidrule(lr){2-4}
 & $\mathbf{Z}^0 = \emptyset$ & $\mathbf{Z}^1 = \{z_1\}$ & $\mathbf{Z}^{12} = \{z_1, z_2\}$ \\
\midrule
Bernoulli  & 0.0444  & 0.0443 & \textbf{0.0425} \\
RandomMLP  & \textbf{0.9024}  & 0.9588 & 0.9218  \\
\bottomrule
\end{tabular}
\end{table}

\subsection{Transportability with observational-interventional mixtures}\label{sec:transportability_results}

\begin{figure}
\centering
\begin{subfigure}{0.45\textwidth}
    \centering
    \includegraphics[width=\textwidth]{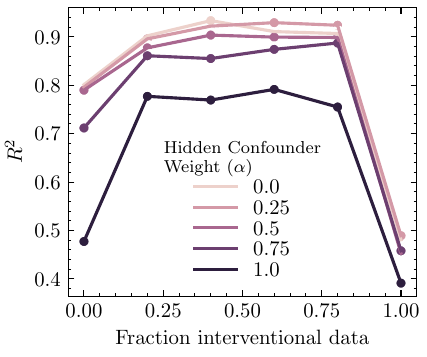}
    \caption{}\label{fig:transportability_r2}
\end{subfigure}
\begin{subfigure}{0.46\textwidth}
    \centering
    \includegraphics[width=\textwidth]{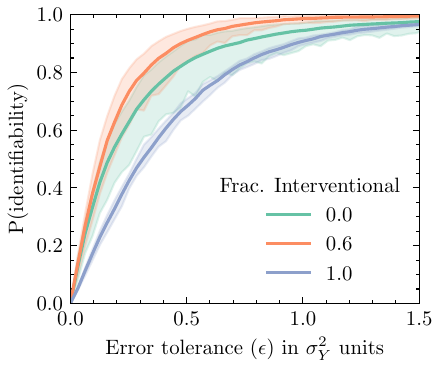}
    \caption{}\label{fig:transportability_curves}
\end{subfigure}
\caption{(a) Meta-model performance in the transportability setting with varying mixtures of observational-interventional data. (b) Identifiability curves for the transportability setting with uncertainty bands across the range of $\alpha$ values.}
\end{figure}

Being able to transport causal effects across populations is a key concern for causal inference in practice. A common use-case for transportability is to be able to take information from a randomized control trial and use it to recover a treatment effect in an observational study. To explore identification in the transportability setting, we consider an example task of being able to identify a causal effect under hidden confounding (\Cref{fig:hidden_confounder_dag}) with access to varying amounts of interventional data, where the source distribution of treatment $T$ in the interventional data \emph{does not match} the distribution in the observational data. Specifically, we consider the following prior over \glspl{scm} shown in \Cref{eq:transportability_equations}.
\begin{equation}\label{eq:transportability_equations}
\begin{aligned}
U_U, U_T, U_Y &\sim \mathcal{N}(0, 1)\\
U &= U_U
\end{aligned}
\quad \quad
\begin{aligned}
T &= f_{t}(\alpha U + U_T\sqrt{1 - \alpha^2})\\
Y &= f_{y}(T, \alpha U + U_Y\sqrt{1 - \alpha^2})
\end{aligned}
\end{equation}
with random \glspl{mlp} $f_t,f_y$ and hidden confounder strength parameterized by $\alpha$, such that we have constant variance $\mathbb{V}(\alpha U + U\sqrt{1 - \alpha^2}) = \alpha^2 \mathbb{V}(U) + (1 - \alpha^2) \mathbb{V}(U)= 1$. We consider varying levels of hidden confounder strength as well as varying fractions of interventional vs. observational data available, where the interventional data come from a regime under $\text{do}(T \leftarrow \mathcal{U}(-1, 1))$, but the target treatment effect to identify is the effect of a unit increase $\text{do}(T \leftarrow T + 1)$ relative to the null-intervention. 

\Cref{fig:transportability_r2} shows $R^2$ of the meta-model in the transportability setting as hidden confounder weight $\alpha$ varies from 0 to 1 and the relative fraction of observational:interventional data varies from 0 to 1. Interventional data do indeed help identify the effect, with a sharp increase in performance, especially under stronger confounding. However, \Cref{fig:transportability_r2} also shows a sharp decrease when all data are interventional, to the point where having all interventional data is worse than having none. This demonstrates the nuance of transporting the causal effect from an interventional regime whose treatment distribution does not directly reflect the target. \Cref{fig:transportability_curves} summarizes this result using the identifiability curve for three fractions (0, 0.6, and 1), where uncertainty bands represent the range of P(identifiability) across $\alpha$ values. In summary, computational identifiability has allowed us to glean an empirical answer for identifiability in this mixture setting and to learn something practical about the mix of observational-interventional data needed to maximize our chances of identifiability.

\subsection{Counterfactual computational identification}\label{sec:counterfactual_results}

\begin{figure}
\centering
\begin{subfigure}{0.49\textwidth}
    \centering
    \includegraphics[width=\textwidth]{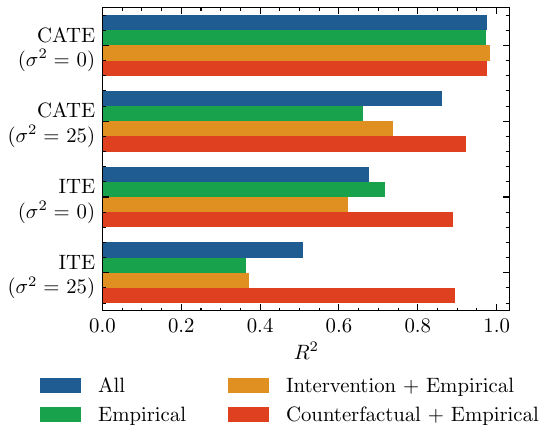}
    \caption{}\label{fig:counterfactual_r2}
\end{subfigure}
\begin{subfigure}{0.42\textwidth}
    \centering
    \includegraphics[width=\textwidth]{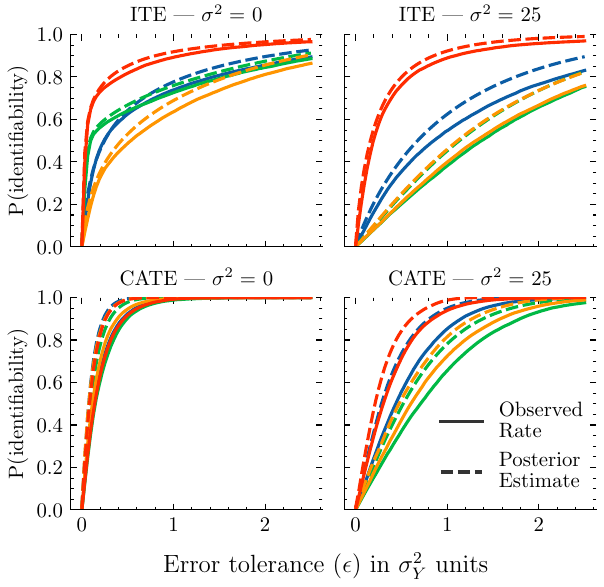}
    \caption{}\label{fig:ite_curves}
\end{subfigure}
\caption{(a) Meta-model performance in the counterfactual setting estimating CATE vs. ITE. (b) Identifiability curves for the counterfactual setting, showing both empirical rates and posterior estimates from \Cref{alg:posterior_p_identifiability}.}
\end{figure}

Counterfactual reasoning is a core component of causal understanding, planning, and decision making. Theoretical counterfactual identification can be particularly challenging, given the need to model combined observational and experimental data as well as introduce shape constraints or otherwise constrain the space of counterfactual distributions in order to reason and transfer information across counterfactual worlds \cite[e.g.,][]{NEURIPS2021_36bedb6e,maiti2025counterfactual}. 

In this experiment, we demonstrate applying computational identifiability to a small finite-sample counterfactual identification question. Specifically, we consider comparing identification of the \gls{cate} to that of the \gls{ite} for the \gls{dag} shown in \Cref{fig:hidden_moderator_dag} and corresponding to \Cref{eq:counterfactual_equations}.
\begin{equation}\label{eq:counterfactual_equations}
\begin{aligned}
\beta_X, \beta_T, \beta_{UT} &\sim \mathcal{U}(-3, 3)\\ \beta_{XT} &\sim \mathcal{U}(-1, 1)\\
X &\sim \mathcal{U}(-2, 2)
\end{aligned}
\quad \quad
\begin{aligned}
T &\sim \text{Bern}(0.5)\\
\epsilon &\sim \mathcal{N}(0, \sigma^2)\\
Y &= \beta_X X + \beta_T T + \beta_{XT} X T + \beta_{UT} U T + \epsilon
\end{aligned}
\end{equation}
In this setting, there is no unobserved confounding, but we have an unobserved variable $U$ that moderates the treatment effect. This makes the \gls{cate}, which is an expected average treatment effect conditional on $X$, distinct from the \gls{ite}, which is instead a unit-specific treatment effect depending also the value of $U$. There is also noise $\epsilon$ on the outcome with variance $\sigma^2$. We consider a mix of empirical, interventional, and counterfactual data, where both interventions and counterfactuals correspond to $\text{do}(T \leftarrow (1 - T))$, providing a `counterfactual' (a corresponding unit with the opposite treatment) for each observed data point. Interventional pairs resample $U$ and resample noise, but condition on $X$, while counterfactual pairs perform abduction (and thus resample neither $U$ nor noise). 

\Cref{fig:counterfactual_r2} shows test $R^2$ after meta-training in 16 different settings with $n=100$ observations. We consider 100\% empirical data (Empirical), 50\% empirical data with interventional pairs (Interventional + Empirical), 50\% empirical data with counterfactual pairs (Counterfactual + Empirical), and an equal mix of all three types (All). We meta-train on each mixture for both the \gls{cate} and the \gls{ite} in a noiseless setting ($\sigma^2=0$) as well as a noisy setting ($\sigma^2=25$). \Cref{fig:ite_curves} shows corresponding P(identifiability) curves for each setting -- this time showing both observed empirical rates (solid lines) and posterior estimates from \Cref{alg:posterior_p_identifiability} (dashed lines). 

Several takeaways about computational identifiability for the \gls{cate} and \gls{ite} are evident here. First, computational identification of the \gls{cate} works in all settings, requiring neither interventional nor counterfactual data. Second, both counterfactual and interventional data can, however, help to recover the \gls{cate} when the outcome is noisy. This is in line with variance reduction from, e.g., matched pairs and/or common random numbers. Third, unlike the \gls{cate}, computational identification of the \gls{ite} requires counterfactual data, and --- as we might expect from the definition of abduction --- works equally well regardless of outcome noise. Finally, the interventional data (Interventional + Empirical) \emph{do not help} identify the \gls{ite} in this case, and if there is no noise on the outcome, \emph{they can instead make things worse}, even under the same interventional regime.

\textbf{Non-monotonic and architecture-specific identifiability.} The counterfactual case above --- and in particular two negative results --- allow us to demonstrate an important point about computational identifiability and its conditional nature. They can be summarized intuitively as `conditional really does mean conditional.' Two parameters of our prior and hypothesis space demonstrate this point well in the \gls{ite} case --- dataset size $N$ and our choice of neural network architecture.

\begin{minipage}[t]{0.375\textwidth}
  \centering\raisebox{\dimexpr \topskip-\height}{%
  \includegraphics[width=\textwidth]{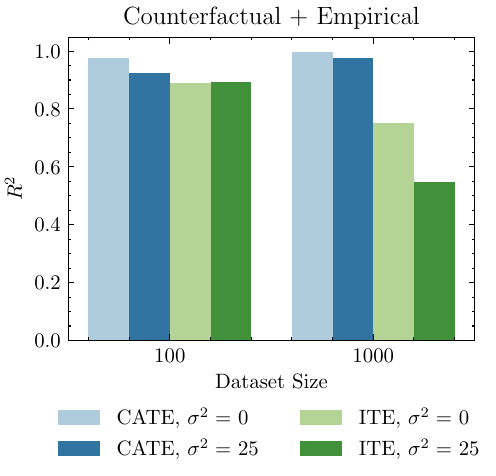}}
  \captionof{figure}{Computational identifiability can have a non-monotonic relationship with dataset size}\label{fig:counterfactual_N}
\end{minipage}\hfill
\begin{minipage}[t]{0.6\textwidth}
  \Cref{fig:counterfactual_N} shows that when we increase dataset size, estimation of the \gls{cate} remains the same or improves, but \gls{ite} estimation can be substantially worse. This demonstrates how computational identifiability can have non-monotonic or surprising relationships with things like increasing dataset size that contradict what we might normally expect from theoretical identifiability. A similar result occurs with architecture choice, in this case whether or not the meta-model processes query points and context sets via cross-attention and set attention (as in the Q-TNP) or via concatenation and mean-pooling (as in the Q-CNP). Taking the (Counterfactual + Empirical), $\sigma^2=25, n=100$ case as an example, the Q-CNP was only able to achieve $R^2=0.4701$, while the Q-TNP was able to achieve $R^2=0.8931$. In other words, in this experiment, only the latter architecture was able to learn to perform the matching of individual datapoints that recovering the \gls{ite} required.
  
\end{minipage}

This property of being conditional on an empirical search --- and its chosen hyperparameters --- is a core component of computational identifiability. The notion itself is fundamentally about attaching a practical, hypothesis-constrained search process to the concept of identifiability.

\section{Conclusion and future work}
Identifiability is not a notion specific only to causal inference, and there is potential to expand the set of definitions we have written in \Cref{sec:computational_identifiability} to work beyond the causal inference settings to other settings, like survival analysis~\citep{ranganath2016deep, miscouridou2018deep} and missing data~\citep{little2019statistical}, where assumptions and analysis drive traditional theoretical identifiability. 
Another important future direction is to study computational identifiability in high dimensional settings, which we have not considered here. A key benefit of computational identifiability, as we have defined it, is that it makes the notion of identifiability \emph{actionable}. Through additional optimization of the computational search process itself, there is also significant potential to further make such a search for identifiability systematic or automated.

\section{Acknowledgments}

This work was partly supported by the NIH/NHLBI Award R01HL148248, NSF Award 1922658 NRT-HDR: FUTURE Foundations, Translation, and Responsibility for Data Science, NSF CAREER Award 2145542, ONR N00014-23-1-2634, NIH R01CA296388, NSF 2404476, Optum and Apple. This work was also supported by IITP with a grant funded by the MSIT of the Republic of Korea in connection with the Global AI Frontier Lab International Collaborative Research.

\bibliographystyle{plainnat}
\bibliography{references}


\appendix

\section{Further generalizing causal meta-prediction frameworks}\label[appendix]{sec:generalizing_other_works}

\begin{figure}[h]
    \centering
    \includegraphics[width=\textwidth]{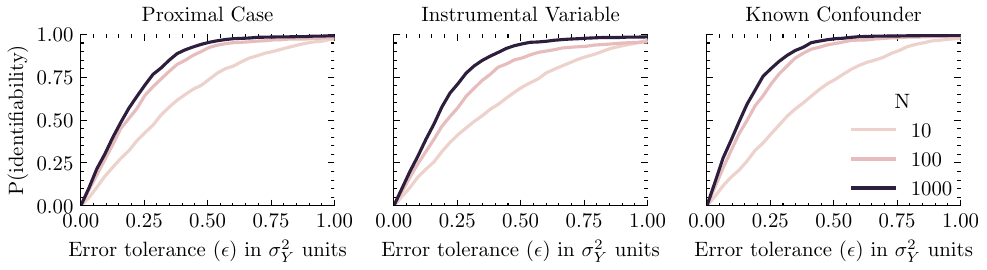}
    \caption{Identifiability curves from meta-models trained on each of the cases in \Cref{eq:proxy_instrument_confounder_setting}, as a function of varying dataset size $N$.}\label{fig:proxy_instrument_confounder}
\end{figure}

Works such as \cite{bynum2025bbci} have proposed frameworks to automate the process of finding estimators via meta-learning. Our abstractions in \Cref{sec:computational_identifiability} further generalize this process, taking the meta-prediction of a causal estimator as one piece of a larger pipeline. To provide some intuition for how meta-prediction of effect estimators fits into our pipeline, we illustrate in this section how experiments from \cite{bynum2025bbci} are equivalent to a special case of our framework and notation. 

\citet[Sec. 3]{bynum2025bbci} introduce the following data generating process in \Cref{eq:proxy_instrument_confounder_setting} to demonstrate meta-learning causal effect estimators in three different settings: a proximal causal inference case (\Cref{fig:proximal_dag}), an instrumental variable case (\Cref{fig:instrument_dag}), and a known confounder case (\Cref{fig:known_confounder_dag}). 
\begin{equation}\label{eq:proxy_instrument_confounder_setting}
    \begin{aligned}
        \delta_1, \delta_2, \delta_3 &\sim \mathcal{U}(0, 1)\\
        \gamma_t, \beta_t, \beta_y &\sim \mathcal{U}(-1, 1)\\
        \gamma_x, \beta_x &\sim \mathcal{U}(-2, -1) \cup \mathcal{U}(1, 2)
    \end{aligned}
    \quad
    \begin{aligned}
        U_Y, Z, X &\sim \mathcal{N}(0, 1)\\
        W_1 &\sim \mathcal{N}(X, \delta_1)\\
        W_2 &\sim \mathcal{N}(X, \delta_2)\\
    \end{aligned}
    \quad \quad \quad 
    \begin{aligned}
        T &= \gamma_x X + \gamma_t Z\\
        Y &= \beta_x X + \beta_t T + \beta_y U_Y
    \end{aligned}
\end{equation}
In the proxy case, $W_1, W_2, T, Y$ are observed; in the instrument case, $Z,T,Y$ are observed; and in the known confounder case, $X,T,Y$ are observed.

In our framework, we can represent these cases via a simple causal mixture distribution, where most of the terms are set to zero ($\alpha^{emp}_{ij} = 0, \alpha^{cf}_{ijk} = 0, \bbId = \emptyset$), and we have $\alpha^{int}_{0} = 1$ for the sole element of $\bbI = \{\emptyset\}$ --- the null intervention. This leaves us with the following for $p^{\fC}_{\bO, mix}$ from \Cref{def:causal_mixture}:
\begin{align*}
p^{\fC}_{\bO, mix}(\bo) &= \sum_{I_j \in \bbId} \sum_{\bo_i \in \cD_j} \left( \alpha^{emp}_{ij} \delta(\bo - \bo_i) + \sum_{I_k \in \bbI} \alpha^{cf}_{ijk} p_{\bO}^{\fC^{\doIj}|\bO=\bo_i; \doIk}(\bo) \right)\\
& \quad \quad + \sum_{I_k \in \bbI} \alpha^{int}_k p_{\bO}^{\fC; \doIk}(\bo)\\
&=\sum_{I_k \in \bbI} \alpha^{int}_k p_{\bO}^{\fC; \doIk}(\bo)\\
&= p^{\fC;\doEmpty}_{\bO}(\bo)\\
&=p^{\fC}_{\bO}(\bo)
\end{align*}
or simply in this case, the observational distribution. In each case, we would have $\bO = \{W_1, W_2, T, Y\}$,  $\bO = \{Z,T,Y\}$, and $\bO = \{X,T,Y\}$, respectively.

We then define the following joint causal-query-mixture distribution according to \Cref{def:joint_causal_query_mixture}, for prior over \glspl{scm} $\pi$ defined by \Cref{eq:proxy_instrument_confounder_setting}, and using the PATE as our causal query for intervention $\text{do}(T \leftarrow T + 1)$ relative to the null-intervention $\text{do}(\emptyset)$:
\begin{align*}
p_{joint}(\bo, \bx, \theta) &= \ECpi{p^{\fC}_{\bO, mix}(\bo) \cdot p_{query}(\bx \mid \fC, \bo) \cdot \delta\left(\theta - \ThetaF{\fC, \bx} \right)}\\
&= \ECpi{p^{\fC}_{\bO}(\bo) \cdot \delta\left(\theta - \Theta_{\text{PATE}}\left[\fC, \cdot \right]\right)}\\
&= \ECpi{p^{\fC}_{\bO}(\bo) \cdot \delta\left(\theta -\bbE_T\left[\mathbb{E}_{p_{Y}^{\fC; \text{do}(T \leftarrow T + 1)}}[Y] - \mathbb{E}_{p_{Y}^{\fC; \text{do}(\emptyset)}}[Y]\right]\right)}
\end{align*}
where query points $\bx$ are not needed, since the PATE is a population-level query (see \Cref{def:causal_query} and the discussion thereafter). 

Having defined a joint causal-query-mixture from which to stream training data $\{(\bo, \theta)\}$, we then meta-train a Q-CNP (see \Cref{sec:architectures}) to perform effect estimation in each of the three cases and compute the empirical probability of identifiability over a new test set sampled from the same distribution.

\Cref{fig:proxy_instrument_confounder} shows the results of meta-training in each of these settings and the resultant identifiability curves as a function of one possible parameter of the computational search: dataset size $N$. We see (1) that meta-training is indeed successful in each of the cases and (2) that as dataset size increases, so too does the probability of identifiability. In this example, we see an illustration of how computational identifiability can allow us to concretely explore identification even with finite sample sizes as small as $N=100$ or $N=10$.

\section{Experiment details}\label[appendix]{sec:experiment_details} 

This section details the experiment architectures, training regimes, and hyperparameters for all of the experiments in this paper. In order to more easily understand how our unified notation in \Cref{sec:computational_identifiability} maps onto the process of meta-training transformers to learn causal estimators, we provide a step-by-step illustration of this mapping to a special case in \Cref{sec:generalizing_other_works}. Additionally, full descriptions of the Q-CNP and Q-TNP architectures can be found in \Cref{sec:architectures}. While several experiments train multiple models, each individual meta-training instance used at most 1 GPU with at most 32GB memory for a compute time of up to 5 hours. Experiments were run on a mix of Intel Xeon Platinum 8592+ 64C CPUs, NVIDIA H200 GPUs, NVIDIA L40S GPUs, and an M5 MacBook Pro.

\begin{table}[H]
\caption{Table of meta-model architectures, training details (dataset size, number of query points, training regime (e.g., offline with a fixed number of examples or online streaming with infinite examples), and size of the test set), as well as hyperparameters (max epochs $e$, embedding size $d$, layers $m$, attention heads $a$, inducing points $i$, and number of quantiles sampled per batch $q$) for each experiment in the paper.}
\label{tab:meta_model_lookup}
\resizebox{\textwidth}{!}{
\centering
\begin{tabular}{cccccccccccc}
\toprule
\multirow{2}{*}{\textbf{Experiment}} & \multirow{2}{*}{\textbf{\shortstack{Meta-model\\Architecture}}} & 
\multirow{2}{*}{\textbf{\shortstack{Dataset\\Size}}} &
\multirow{2}{*}{\textbf{\shortstack{Query\\Points}}} & \multirow{2}{*}{\textbf{\shortstack{Train Mode\\(Train Size)}}} & \multirow{2}{*}{\textbf{\shortstack{Test\\Size}}} & \multicolumn{6}{c}{\textbf{Hyperparameters}}\\\cmidrule(lr){7-12}
&&&&&& $e$ & $d$ & $m$ & $a$ & $i$ & $q$\\\midrule

\Cref{tab:oset_table} & Q-CNP & 1000 & 30 & Online & 1000 & 500 & 128 & 2 & - & - & 32\\

\Cref{fig:transportability_r2,fig:transportability_curves} & Q-CNP & 1000 & 30 & Offline (10k) & 1000 & 500 & 128 & 8 & - & - & 32\\

\Cref{fig:counterfactual_r2,fig:ite_curves} & Q-TNP & 100 & 30 & Offline (10k) & 1000 & 500 & 128 & 2 & 4 & 32 & 32\\

ITE Architecture & Q-CNP/Q-TNP & 100 & 30 & Offline (10k) & 1000 & 500 & 128 & 2 & -/4 & -/32 & 32\\

\Cref{fig:counterfactual_N} & Q-TNP & 100/1000 & 30 & Offline (10k) & 1000 & 500 & 128 & 2 & 4 & 32 & 32\\

\Cref{fig:proxy_instrument_confounder} & Q-CNP & 10/100/1000 & - & Offline (10k) & 1000 & 500 & 128 & 8 & - & - & 32\\

\Cref{fig:classical_curves} & - & 10/100/1000 & - & - & 1000 & - & - & - & - & - & -\\

\bottomrule
\end{tabular}}
\end{table} 

\section{Computational identifiability with per-dataset estimation methods}\label[appendix]{sec:per_dataset_methods}

\begin{figure}[H]
    \centering
    \includegraphics[width=\textwidth]{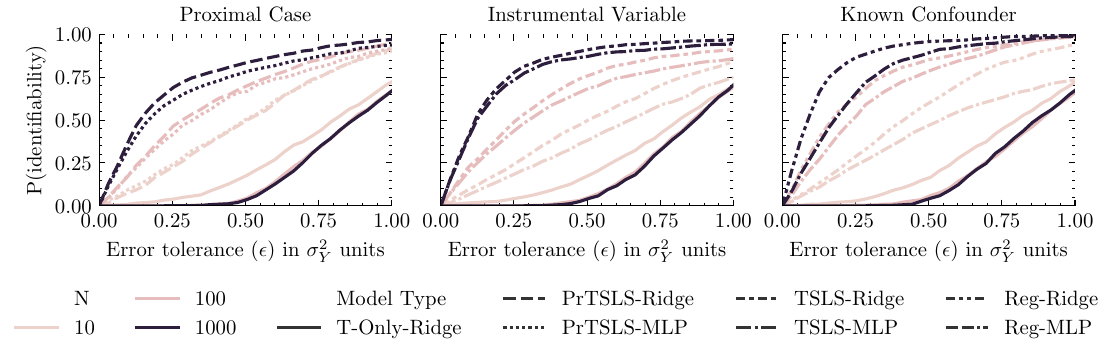}
    \caption{Identifiability curves from per-dataset estimation methods instead of meta-trained models, for each of the cases in \Cref{eq:proxy_instrument_confounder_setting}, as a function of varying dataset size $N$.}\label{fig:classical_curves}
\end{figure}

Using the example in \Cref{sec:generalizing_other_works}, we can demonstrate a useful point about computational identifiability; namely, that \emph{the notion itself does not have to require meta-training}, and we can incorporate existing per-dataset algorithms into a computational identifiability search. \Cref{fig:classical_curves} shows exactly this, showing the probability of identifiability computed by empirically evaluating each per-dataset method described in \citet[Sec. 3]{bynum2025bbci}: proximal two-stage least squares with ridge regression (PrTSLS-Ridge), proximal two-stage least squares with an MLP first stage and ridge regression second stage (PrTSLS-MLP), two-stage least squares with ridge regression (TSLS-Ridge), two-stage least squares with an MLP first stage and ridge regression second stage (TSLS-MLP), regular ridge regression (Reg-Ridge), regression with an MLP (Reg-MLP), and finally ridge regression using treatment only (T-Only-Ridge). 

This evaluation is done on each of the same datasets the meta-model in \Cref{fig:proxy_instrument_confounder} was evaluated on. However, comparing the superior performance of the meta-model to that of the baselines (especially those that perform poorly) illustrates the utility of having a meta-model even in settings with known algorithms, provided the meta-model is able to perform well. The story about identifiability in the meta-model search is, for example, more optimistic than in the known-algorithm search, especially for small dataset sizes.


\end{document}